\documentclass[conference]{IEEEtran}
\IEEEoverridecommandlockouts

\usepackage{cite}
\usepackage{amsmath,amssymb,amsfonts}
\usepackage{algorithmic}
\usepackage{graphicx}
\usepackage{textcomp}
\usepackage{xcolor}
\usepackage{hyperref}
\usepackage{orcidlink}
\usepackage[normalem]{ulem}
\usepackage{comment}
\usepackage{algorithm2e}

\def\BibTeX{{\rm B\kern-.05em{\sc i\kern-.025em b}\kern-.08em
    T\kern-.1667em\lower.7ex\hbox{E}\kern-.125emX}}

\begin{document}
\small	

\title{CNC-TP: Classifier Nominal Concept based on Top-Pertinent attributes\

}

\author{
    \IEEEauthorblockN{Yasmine Souissi\IEEEauthorrefmark{1},
    Fabrice Boissier\,\orcidlink{0000-0002-0067-6524}\IEEEauthorrefmark{1}\IEEEauthorrefmark{2},
    Nidà Meddouri\,\orcidlink{0000-0002-7815-630X}\IEEEauthorrefmark{1}}
    \IEEEauthorblockA{\IEEEauthorrefmark{1}\textit{LRE} - \textit{EPITA}
    \\\{yasmine.souissi, fabrice.boissier, nida.meddouri\}@epita.fr}
    \IEEEauthorblockA{\IEEEauthorrefmark{2}\textit{CRI} - \textit{Université Paris 1 Panthéon - Sorbonne}
    }
}

\maketitle

\begin{abstract}
Knowledge Discovery in Databases (KDD) aims to exploit the vast amounts of data generated daily across various domains of computer applications. 
Its objective is to extract hidden and meaningful knowledge from datasets through a structured process comprising several key steps: \textit{data selection}, \textit{preprocessing}, \textit{transformation}, \textit{data mining}, and \textit{visualization}.
Among the core data mining techniques are \textit{classification} and \textit{clustering}. 
Classification involves predicting the class of new instances using a classifier trained on labeled data. 
Several approaches have been proposed in the literature, including \textit{Decision Tree Induction}, \textit{Bayesian classifiers}, \textit{Nearest Neighbor search}, \textit{Neural Networks}, \textit{Support Vector Machines}, and \textit{Formal Concept Analysis} (FCA).
The last one is recognized as an effective approach for interpretable and explainable learning. 
It is grounded in the mathematical structure of the \textit{concept lattice}, which enables the generation of formal concepts and the discovery of hidden relationships among them.
In this paper, we present a state-of-the-art review of FCA-based classifiers. 
We explore various methods for computing closure operators from nominal data and introduce a \textit{novel approach} for constructing a partial concept lattice that focuses on the most relevant concepts. 
Experimental results are provided to demonstrate the efficiency of the proposed method.
\end{abstract}



\section{Introduction}
\label{section:Introduction}

\par Classification based on Formal Concept Analysis (FCA)~\cite{ganter2012formal} is a machine learning approach that leverages rule induction to construct classifiers. 
It is grounded in the mathematical framework of formal contexts, Galois connections, and concept lattices, which are used to derive classification rules from data.

\par Several FCA-based classification methods have been proposed recently, including \textit{CNC (Classifier Nominal Concept)}~\cite{MeddouriPAKDD2012}, \textit{CpNC\_CORV (Classifier pertinent Nominal Concept based on Closure Operator for Relevant-Values)}~\cite{Meddouri2020EfficientClosure}, \textit{DFC (Dagging Formal Concept)}~\cite{Meddouri2021DFC}, \textit{NextPriorityConcept}~\cite{Boukhetta2020SequenceMining}, and \textit{Adapted SAMME Boosting}~\cite{Semenkov2021EnsembleTechniques}. 
Despite their contributions, these methods face several challenges, such as high error and rejection rates, as well as overfitting.

\par Furthermore, the exhaustive construction of all formal concepts can be computationally intensive and often lacks contextual relevance. 
There is also a notable absence of adaptive strategies for selecting the most pertinent concepts~\cite{MeddouriCLA2008}. 
To address these limitations, various enhancements have been explored, including ensemble learning techniques~\cite{MeddouriPAKDD2012, MeddouriKES:2014} and pattern structures~\cite{Semenkov2021EnsembleTechniques}. 
However, the quality of the generated concepts used for classification remains an under-explored aspect.

\par In this work, we propose an enhancement of the CNC method by introducing a novel strategy for selecting relevant attributes and constructing classifiers based on formal concepts. 
Section~\ref{section:FCA_basics} introduces the fundamentals of Formal Concept Analysis. 
Section~\ref{section:state_art} reviews recent FCA-based classification methods. 
Section~\ref{section:proposed_approach} presents our proposed method, CNC-TP. 
Section~\ref{section:experimentations} details the experimental evaluation of our approach. 
Finally, Section~\ref{section:conclusion} concludes the paper with a summary and perspectives for future work.


\section{Formal Concept Analysis}
\label{section:FCA_basics}

\par Formal Concept Analysis (FCA)~\cite{ganter2012formal}\cite{WESTPHAL2022108233}\cite{stumme2005fca} is a mathematical framework rooted in lattice theory and propositional logic. 
It is used to extract conceptual structures from data by identifying and organizing formal concepts into a concept lattice.
FCA operates on a \textit{formal context}, which represents the relationships between a set of objects (instances) and a set of attributes. 
To illustrate this, we draw inspiration from the well-known \textit{weather.symbolic} dataset, which describes meteorological conditions and the decision of whether to play outside.
Table~\ref{table:hal-FCA-weather-nominal-dataset} presents an example of a formal and nominal context, containing the first seven instances of the \textit{weather.symbolic} dataset.

\begin{table}[!ht]
\centering
\caption{Subset of the \textit{weather.symbolic} data-set (7 first instances)}
\label{table:hal-FCA-weather-nominal-dataset}
\begin{tabular}{|c|c|c|c|c|c|}
\hline
ID & Outlook  & Temperature & Humidity & Windy & Play \\
\hline
1 & sunny    & hot         & high     & FALSE & no   \\
2 & sunny    & hot         & high     & TRUE  & no   \\
3 & overcast & hot         & high     & FALSE & yes  \\
4 & rainy    & mild        & high     & FALSE & yes  \\
5 & rainy    & cool        & normal   & FALSE & yes  \\
6 & rainy    & cool        & normal   & TRUE  & no   \\
7 & overcast & cool        & normal   & TRUE  & yes  \\
\hline
\end{tabular}
\end{table}

\par The symbolic (nominal) dataset is transformed into a binary \textit{formal context}, where each attribute–value pair is represented as a distinct binary feature. The resulting binary context is presented in Table~\ref{table:half-FCA-weather-nominal-context}.
Table~\ref{table:half-FCA-weather-mapping} provides the mapping between each binary attribute $a_i$ and its corresponding attribute–value pair from the original symbolic dataset.

\begin{table}
\centering
\caption{Example of formal and binary context from Table \ref{table:hal-FCA-weather-nominal-dataset} .}
\label{table:half-FCA-weather-nominal-context}
\begin{tabular}{|c|c|c|c|c|c|c|c|c|c|}
\hline
ID & $a_1$ & $a_2$ & $a_3$ & $a_4$ & $a_5$ & $a_6$ & $a_7$ & $a_8$ & $a_9$ \\
\hline
$i_1$ & 1 & 0 & 0 & 1 & 0 & 0 & 1 & 0 & 0 \\
$i_2$ & 1 & 0 & 0 & 1 & 0 & 0 & 1 & 0 & 1 \\
$i_3$ & 0 & 1 & 0 & 1 & 0 & 0 & 1 & 0 & 0 \\
$i_4$ & 0 & 0 & 1 & 0 & 1 & 0 & 1 & 0 & 0 \\
$i_5$ & 0 & 0 & 1 & 0 & 0 & 1 & 0 & 1 & 0 \\
$i_6$ & 0 & 0 & 1 & 0 & 0 & 1 & 0 & 1 & 1 \\
$i_7$ & 0 & 1 & 0 & 0 & 0 & 1 & 0 & 1 & 1 \\
\hline
\end{tabular}
\end{table}

\begin{table}
\centering
\caption{Signification of binary attributes from previous binary context}
\label{table:half-FCA-weather-mapping}
\begin{tabular}{|c|l|}
\hline
Binary Attribute & Description \\
\hline
$a_1$ & Outlook = sunny \\
$a_2$ & Outlook = overcast \\
$a_3$ & Outlook = rainy \\
$a_4$ & Temperature = hot \\
$a_5$ & Temperature = mild \\
$a_6$ & Temperature = cool \\
$a_7$ & Humidity = high \\
$a_8$ & Humidity = normal \\
$a_9$ & Windy = TRUE \\
\hline
\end{tabular}
\end{table}

\par We define the \textit{formal context} as a triplet \( \langle I, A, R \rangle \), where:
\begin{itemize}
    \item \( I = \{i_1, i_2, \dots, i_7\} \) is the set of objects (instances),
    \item \( A = \{a_1, a_2, \dots, a_9\} \) is the set of binary attributes (e.g., $a_1$: outlook = sunny, $a_4$: temperature = hot, ...),
    \item \( R \subseteq I \times A \) is the binary relation such that \( R(i_k, a_l) = 1 \) if instance \( i_k \) possesses attribute \( a_l \).
\end{itemize}

\par Two operators define the \textit{Galois connection}:
\begin{itemize}
    \item \( \varphi(X) = \{ a \in A \mid \forall i \in X, (i, a) \in R \} \): attributes common to all objects in \( X \),
    \item \( \delta(Y) = \{ i \in I \mid \forall a \in Y, (i, a) \in R \} \): objects sharing all attributes in \( Y \).
\end{itemize}

\par These operators allow us to compute the \textit{closures}:
\[ X" = \delta(\varphi(X)), \quad Y" = \varphi(\delta(Y)) \]

\par A formal concept is a pair \( (X, Y) \) such that \( \varphi(X) = Y \) and \( \delta(Y) = X \). 

\par For example, if $X = \{i_1, i_2\}$ then: 
\[ \varphi(X) = \{a_1, a_4, a_7\}, \quad \delta(\{a_1, a_4, a_7\}) = \{i_1, i_2\}\]
So, \( (\{i_1, i_2\}, \{a_1, a_4, a_7\}) \) is a formal concept.

\bigskip

\par
\begin{figure}[!ht]
\centering
\includegraphics[width=1.0\linewidth]{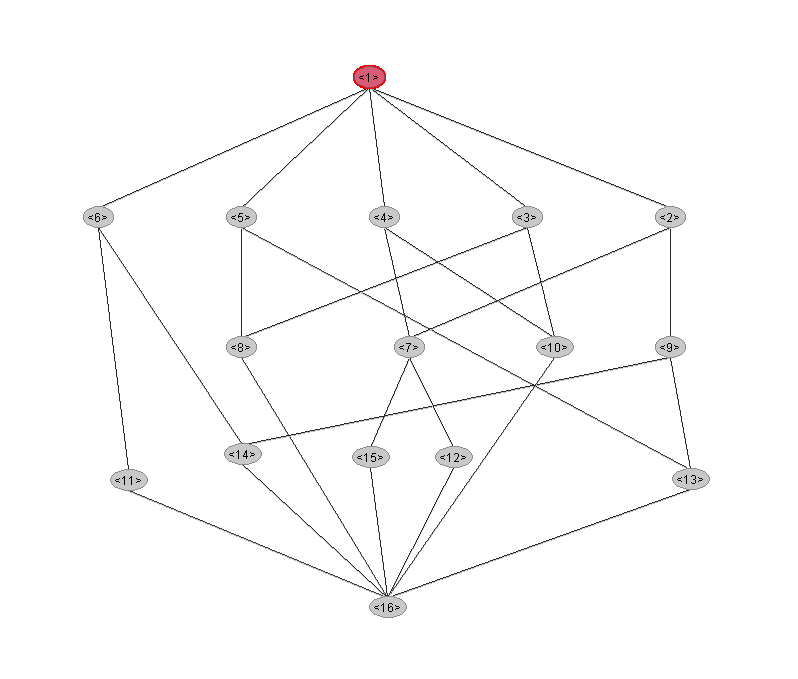}
\caption{Concept lattice generated from the previous formal context using Galicia}
\label{fig:lattice}
\end{figure}

 These concepts can be organized into a lattice structure, where each node represents a concept and edges define the partial order based on set inclusion. 
This hierarchical structure reveals how concepts are related through shared attributes and objects, and highlights generalization–specialization relationships.
Figure~\ref{fig:lattice} shows the concept lattice derived from the \textit{weather.symbolic} formal context using the Galicia tool\footnote{\url{http://www.iro.umontreal.ca/~galicia/}} \cite{Grosser2005}.
This lattice serves as a foundation for many FCA-based applications, including rule extraction and classification.

\section{State of the Art for Classification Based on Formal Concept Analysis}
\label{section:state_art}

\par Formal Concept Analysis (FCA)~\cite{ganter2012formal}\cite{WESTPHAL2022108233}\cite{stumme2005fca} has attracted growing interest across various domains due to its ability to extract structured knowledge from binary and nominal data. This section reviews significant contributions that demonstrate how FCA has been applied to real-world classification problems. We focus particularly on FCA-based classifiers designed to handle nominal data, highlighting their strengths, limitations, and recent improvements. These works underscore both the versatility of FCA in data analysis and the ongoing efforts to enhance its scalability and interpretability for complex datasets.

\par In~\cite{Semenkov2021EnsembleTechniques}, the authors propose a method for generating formal concepts from empirical data by combining classical (\textit{a posteriori}) approaches, such as traditional FCA, with non-classical (\textit{a priori}) methodologies like the System of Measured Properties (SMP). The study also addresses the challenge of incomplete and inconsistent data using multi-valued logic. The authors conclude that both methodologies are essential for deriving meaningful formal concepts, and emphasize the importance of normalizing the formal context when dealing with multi-valued logic.

\par The work in~\cite{Meddouri2020EfficientClosure} introduces a novel approach for extracting pertinent concepts from nominal data using closure operators derived from Galois connections. The authors propose and evaluate four classification methods: \textbf{CpNC\_COMV} (Closure Operator for Multi-Values), \textbf{CpNC\_CORV} (Closure Operator for Relevant-Values), \textbf{CaNC\_COMV}, and \textbf{CaNC\_CORV}. Experimental results demonstrate that \textbf{CpNC\_CORV} outperforms the others in terms of classification accuracy.

\par In~\cite{Meddouri2021DFC}, the authors propose a method for constructing only a subset of the concept lattice, focusing on the most relevant concepts. They introduce the \textbf{DNC} (Dagging Nominal Concept) classifier, which leverages the \textit{Dagging} technique to create ensembles of FCA-based classifiers. The study shows that parallel learning significantly improves performance compared to single classifiers. While this approach reduces computational complexity, it may also lead to information loss by omitting less prominent concepts, potentially affecting classification precision.

\par The \textit{NextPriorityConcept} algorithm, introduced in~\cite{Demko2020NextPriorityConcept}, uses priority queues and filtering strategies to compute concepts efficiently. It treats all data uniformly by converting them into logical predicates, making it format-independent. However, the algorithm may face scalability issues when applied to large or highly complex datasets due to exponential growth in computational complexity.

\par In another contribution from~\cite{Semenkov2021EnsembleTechniques}, the authors explore ensemble methods to enhance lazy classification using pattern structures. While pattern structures offer interpretable models, they often underperform compared to traditional ensemble techniques. To address this, the authors adapt the SAMME boosting algorithm to pattern structures and investigate various aggregation functions and weighting schemes. Their work demonstrates that it is possible to improve predictive accuracy without compromising interpretability.

\par Finally, in~\cite{Boissier2024FCA}, FCA is applied to corpus visualization as an alternative to traditional topic modeling techniques. The study shows that FCA can effectively visualize keywords and document relevance, offering a promising alternative to Latent Dirichlet Allocation (LDA)~\cite{blei2003latent} for topic modeling and relevance analysis.

\par Despite the diversity and progress of FCA-based classification methods, several common limitations persist across the literature. 
First, many approaches suffer from high computational complexity due to the exhaustive generation of formal concepts, which becomes impractical for large or complex datasets. 
While some methods attempt to reduce this burden by constructing only a subset of the concept lattice, this often leads to a loss of potentially useful information, thereby affecting classification accuracy. 
Second, most existing techniques lack contextual adaptability, as they do not dynamically select the most relevant concepts based on the classification task. 
Third, handling nominal and multi-valued data remains a challenge, particularly during the transformation into binary contexts, which can introduce redundancy or information loss. 
Additionally, several methods exhibit high error or rejection rates, especially when the generated concepts are overly specific or insufficiently discriminative. 
Overfitting is another recurring issue, particularly in models that generate a large number of rules or concepts. 
Furthermore, few studies explicitly evaluate the quality or discriminative power of the concepts used for classification. 
Finally, although ensemble methods such as \textit{Dagging} and \textit{SAMME Boosting} have improved performance, they can compromise model interpretability—one of the core advantages of FCA.


\section{Proposed Approach: CNC-TP}
\label{section:proposed_approach}

\par The previously reviewed studies demonstrate the adaptability and effectiveness of FCA in addressing various data types and classification challenges. 
From optimizing concept extraction to enhancing classifier performance through ensemble techniques and specialized measures, FCA continues to evolve as a powerful analytical framework. 
However, challenges remain—particularly in balancing computational complexity and classification precision when dealing with large-scale or multi-valued datasets. 
These insights motivate the development of more refined FCA-based classification methods, as proposed in this work.

\par Our proposed approach builds upon the \textit{Classifier Nominal Concept} (\textbf{CNC})~\cite{Meddouri2020EfficientClosure}, a classification method that leverages nominal attributes to induce decision rules. 
Starting from a nominal (multi-valued) context, CNC extracts the most pertinent nominal concepts by computing Galois connections (closure operators) on attributes that maximize \textit{Information Gain}, thereby improving data separability. 
The primary objective of CNC is to reduce training time and complexity while ensuring the interpretability and explainability of the resulting model.

\par In this work, we introduce \textbf{CNC-TP} (CNC based on Top Pertinent attributes), a classification approach that enhances CNC by selecting the most informative attributes using the \textit{Gain Ratio} measure. 
Unlike traditional FCA-based classifiers that consider all attributes or rely on fixed thresholds, CNC-TP dynamically selects a top-ranked subset of attributes based on their relevance. 
This strategy improves both the interpretability and performance of the classifier by reducing noise and focusing on the most discriminative patterns. 
The approach follows a structured pipeline: attribute ranking, formal concept extraction, and rule-based classification.

\par The core steps of \textbf{CNC-TP} involve selecting the most relevant attributes according to the \textit{Gain Ratio} pertinence measure, followed by computing the \textbf{Galois Connection} to generate only the formal concepts derived from these attributes. Rather than retaining all attributes above a fixed threshold, CNC-TP selects the top-$p\%$ of attributes ranked by Gain Ratio, where $p$ is either user-defined or automatically determined by the classifier.

\subsection{Classifier Construction}
\label{subsection:Classifier_Construction}

\begin{algorithm} \small \footnotesize	
{
\KwIn{
Training set $O = \{(o_1, y_1), \dots, (o_n, y_n)\}$ where $y_i \in \mathcal{K}$ (class labels) and described by m attributes $A = \{a_1, \dots, a_n\}$.\\
A threshold $p \in [0, 100]$.
}
\KwOut{
A set of classification rules with weights.
}
\Begin{
    \ForEach{attribute $a_i$ in $A$}{
        Compute the \textbf{Gain Ratio} of $a_i$\;
    }
    Sort all attributes in descending order of Gain Ratio\;
    Select the top-$p\%$ attributes based on the $p$ threshold \;
    
    \ForEach{selected attribute}{
        Apply a chosen strategy to compute the Galois connection\;
        \tcp{Use either all values or majority values}
        Extract formal concepts from the attribute-value pairs\;
    }

    \ForEach{formal concept}{
        Generate a classification rule $R$\;
        Initialize $N_{\text{covered}} \leftarrow 0$\;
        Let $N_{\text{total}} \leftarrow$ number of instances in $O$\;
        \ForEach{instance $o_i$ in $O$}{
            \If{$R(o_i)$ is satisfied}{
                $N_{\text{covered}} \leftarrow N_{\text{covered}} + 1$\;
            }
        }
        Compute the rule weight:
        \[
        \text{Weight}(R) = \frac{N_{\text{covered}}}{N_{\text{total}}}
        \]
        Store $(R, \text{Weight}(R))$ in the rule set\;
    }

    Use a voting method (e.g., weighted majority vote) to classify new instances based on the set of rules\;
}
}
\caption{\textsc{Classifier Nominal Concept based on Top-Pertinent attributes Algorithm}}
\label{Algo_CNC_TP}
\end{algorithm}

\par Algorithm~\ref{Algo_CNC_TP} outlines the construction process of the CNC-TP classifier. 
It begins by selecting the most relevant attributes using the Gain Ratio measure, then derives formal concepts based on these attributes, generates classification rules, and assigns a weight to each rule based on its coverage. 
Finally, the classifier uses a voting mechanism to predict the class of new instances.

\subsection{Concept Extraction Strategies}
\label{subsection:Concept_Extraction_Strategies}

\par \textbf{CNC-TP} supports two strategies for concept extraction:
\begin{itemize}
    \item \textbf{Top-$p\%$ Attributes with All Values:} Galois connections are computed for all values of the top-ranked attributes.
    \item \textbf{Top-$p\%$ Attributes with Relevant Values:} Only the majority value of each selected attribute is considered for concept generation.
\end{itemize}

\par To identify the most informative attributes, we employ the \textit{Ratio Gain} measure. 
As introduced in~\cite{quinlan1986induction}, the Ratio Gain is designed to mitigate the bias of the traditional \textit{Information Gain}, which tends to favor attributes with a large number of distinct values. 
The Ratio Gain evaluates the usefulness of an attribute by considering both its information gain and the intrinsic information of its value distribution. 
It favors attributes that provide a balanced and meaningful partitioning of the data, penalizing overly specific or fragmented splits. 
This makes it particularly suitable for selecting attributes in FCA-based classification, where interpretability and generalization are key objectives.

\par First, the attributes are sorted in descending order according to their \textit{Gain Ratio (GR)} values.

\subsection{Concepts and Rules Generation}
\label{subsection:Concept_Rules_Generation}

\par After selecting the most relevant attributes, the next step is to compute the Galois connection by identifying the set of instances associated with each attribute value. 
Then, the closure operator is applied to determine additional attributes that are common to all selected instances.

\par In our example, we extract concepts based on the values of the selected attribute \textbf{outlook}. 
For each value:
\begin{itemize}
    \item \textbf{sunny}: covers instances $o_1$ and $o_2$, both labeled \textit{no} \\
    $\Rightarrow$ Concept: $[\{\textit{Outlook = sunny}\}, \{o_1, o_2\}]$
    \item \textbf{overcast}: covers instances $o_3$ and $o_7$, both labeled \textit{yes} \\
    $\Rightarrow$ Concept: $[\{\textit{Outlook = overcast}\}, \{o_3, o_7\}]$
    \item \textbf{rainy}: covers instances $o_4$, $o_5$, and $o_6$; two labeled \textit{yes}, one labeled \textit{no} \\
    $\Rightarrow$ Concept: $[\{\textit{Outlook = rainy}\}, \{o_4, o_5, o_6\}]$ (majority class: \textit{yes})
\end{itemize}

\par At the end of this step, we obtain a set of formal (nominal) concepts, each of which is used to generate a classification rule. Each rule consists of:
\begin{itemize}
    \item \textbf{Premises:} derived from the attribute–value pair(s) defining the concept.
    \item \textbf{Conclusion:} the majority class among the instances covered by the concept.
    \item \textbf{Weight:} the proportion of instances covered by the rule relative to the total dataset.
\end{itemize}

\par The resulting rules from our example are:
\begin{itemize}
    \item \textbf{If} Outlook = sunny \textbf{then} play = no \hfill (coverage: $2/7$)
    \item \textbf{If} Outlook = overcast \textbf{then} play = yes \hfill (coverage: $2/7$)
    \item \textbf{If} Outlook = rainy \textbf{then} play = yes \hfill (coverage: $2/7$)
\end{itemize}

\subsection{Classification by CNC-TP}
\label{subsection:classification_CNC_TP}

\par To classify a new instance, all applicable rules generated during the training phase are identified and evaluated. 
A \textbf{weighted majority voting} strategy is employed, where each rule contributes to the final decision proportionally to its coverage ratio. 
This ensures that rules covering a larger portion of the dataset have a stronger influence on the classification outcome.

\par A distinctive feature of the CNC-TP approach is its ability to \textbf{reject classification} for instances that are not covered by any rule. 
This mechanism prevents unreliable predictions and ensures that only confidently supported decisions are made, thereby enhancing the robustness of the classifier.


\section{Experimentation}
\label{section:experimentations}

\par To evaluate the effectiveness of the proposed \textbf{CNC-TP} method, we conducted a series of experiments comparing its performance with other FCA-based and classical classification approaches across multiple benchmark datasets. 
The goal is to assess its accuracy, interpretability, and robustness under various data conditions.
The CNC-TP method was implemented within the WEKA framework~\cite{Frank:2005}, and the source code is publicly available\footnote{\url{https://gitlab.com/nidameddouri/2025_canc_3.0}} for reproducibility and further experimentation.

\subsection{Experimental Protocol}
\label{subsection:experimental_protocol}

\par In our experimental study, we adopted a 10-fold cross-validation strategy to evaluate the performance and generalization ability of the proposed \textbf{CNC-TP} classifier. 
According to~\cite{berrar2018cross}, cross-validation is a widely used resampling technique that helps assess the robustness of predictive models and mitigate overfitting. 
In $k$-fold cross-validation, the dataset is partitioned into $k$ disjoint subsets (folds) of approximately equal size. 
The model is trained on $k-1$ folds and tested on the remaining fold. 
This process is repeated $k$ times, ensuring that each instance is used exactly once for validation. The final performance is computed as the average across all $k$ iterations.

\par In our case, we used $k = 10$, meaning that the dataset was divided into 10 folds. 
For each iteration, the model was trained on 9 folds and evaluated on the remaining one. 
This setup ensures a reliable and unbiased estimation of the classifier’s predictive performance.

\par To assess the effectiveness of CNC-TP, we employed a comprehensive set of evaluation metrics, covering both classification accuracy and error analysis, as well as probabilistic and ranking-based measures: Accuracy, Percent Incorrect, Percent Unclassified, Precision, Recall, F-Measure, Kappa, AUC-ROC, AUC-PRC, RMSE, MAE.
These metrics provide a multidimensional evaluation of the classifier’s performance, capturing not only its accuracy but also its reliability, robustness, and ability to handle uncertainty and class imbalance.

\subsection{Data Sets}
\label{subsection:data_sets}

\par For our experimental evaluation, we selected 16 datasets from the UCI Machine Learning Repository, covering a wide range of domains and data characteristics. These datasets include both binary and multi-class classification tasks, and feature a mix of nominal, numeric, and symbolic attributes. Table~\ref{table:experimental_Datasets_extended} summarizes the key properties of each dataset, including the number of attributes and instances, class distribution, percentage of missing values, entropy, Gini index, and Palma ratio.

\begin{table*}[!ht]
\centering
\caption{Summary of datasets with missing values, entropy, Gini index and Palma ratio}
\label{table:experimental_Datasets_extended}
\begin{tabular}{|l|c|c|c|c|c|c|c|}
\hline
\textbf{Dataset} & \textbf{Attributes} & \textbf{Instances} & \textbf{Classes} & \textbf{\% Missing} & \textbf{Entropy} & \textbf{Gini Index} & \textbf{Palma Ratio} \\
\hline
Balance-scale & 5 & 625 & 3 & 0 & 1.318 & 0.5692 & 0.855 \\
Breast-cancer & 10 & 277 & 2 & 9 & 0.878 & 0.9412 & 3.519 \\
Contact-lenses & 5 & 24 & 3 & 0 & 1.326 & 0.9082 & 1.667 \\
Credit-rating & 16 & 653 & 2 & 67 & 0.991 & 0.9038 & 1.206 \\
Glass & 10 & 214 & 6 & 0 & 2.177 & 0.7367 & 0.55 \\
Heart-statlog & 14 & 270 & 2 & 0 & 0.991 & 0.4938 & 1.25 \\
Iris & 5 & 150 & 3 & 0 & 1.585 & 0.6667 & 0.5 \\
Iris-2D & 5 & 150 & 3 & 0 & 1.585 & 0.6667 & 0.5 \\
Labor-neg-data & 17 & 57 & 2 & 326 & 0.935 & 0.8642 & inf \\
Lupus & 4 & 87 & 2 & 0 & 0.972 & 0.4809 & 1.486 \\
Pima\_diabetes & 9 & 768 & 2 & 24 & 0.933 & 0.4544 & 1.865 \\
Vowel & 14 & 990 & 11 & 0 & 3.459 & 0.9274 & 0.1 \\
Weather.numeric & 5 & 14 & 2 & 0 & 0.94 & 0.9106 & 1.8 \\
Weather.symbolic & 5 & 14 & 2 & 0 & 0.94 & 0.9106 & 1.8 \\
Wisconsin-breast-cancer & 10 & 683 & 2 & 16 & 0.929 & 0.455 & 1.857 \\
Zoo & 18 & 101 & 7 & 0 & 2.391 & 0.6034 & 0.683 \\
\hline
\end{tabular}
\end{table*}

\par The diversity of these datasets allows for a robust assessment of the proposed \textbf{CNC-TP} method under various conditions. 
For instance, datasets such as \textit{vowel}, \textit{zoo}, and \textit{glass} present multi-class challenges, while others like \textit{lupus}, \textit{heart-statlog}, and \textit{pima\_diabetes} focus on binary classification. 
The presence of missing values in datasets such as \textit{labor-neg-data}, \textit{credit-rating}, and \textit{wisconsin-breast-cancer} tests the classifier’s resilience to incomplete data.

\par The entropy and Gini index values provide insight into the class distribution and impurity of each dataset. 
Higher entropy (e.g., \textit{vowel}: 3.459) indicates more evenly distributed classes, while lower values (e.g., \textit{lupus}: 0.972) suggest class imbalance. 
Similarly, the Gini index reflects the heterogeneity of the data; values close to 1 (e.g., \textit{breast-cancer}: 0.9412) indicate high impurity, which can complicate classification.

\par The Palma ratio, a measure of inequality adapted from economics, is used here to quantify the imbalance in attribute distributions. 
A low Palma ratio (e.g., \textit{vowel}: 0.1) suggests uniform attribute distribution, while high values (e.g., \textit{breast-cancer}: 3.519) indicate strong skewness, which may affect rule generation and classifier performance.

\par Overall, the selected datasets offer a comprehensive testbed for evaluating the adaptability, interpretability, and robustness of the CNC-TP classifier across diverse data scenarios.

\subsection{CNC-TP-AV: Top Pertinent Attributes with All Values}
\label{subsection:CNC-TP-AV}

\par In this experiment, we evaluate the performance of the \textbf{CNC-TP} classifier using the first concept extraction strategy described in Section~\ref{subsection:Concept_Extraction_Strategies}, which selects the top-ranked attributes and considers all their values. 
The threshold for attribute selection is varied from 0.025 (2.5\%) to 1.000 (100\%) to identify the optimal value that yields the best classification performance.

\begin{figure}[!ht]
\centering
\includegraphics[width=\linewidth]{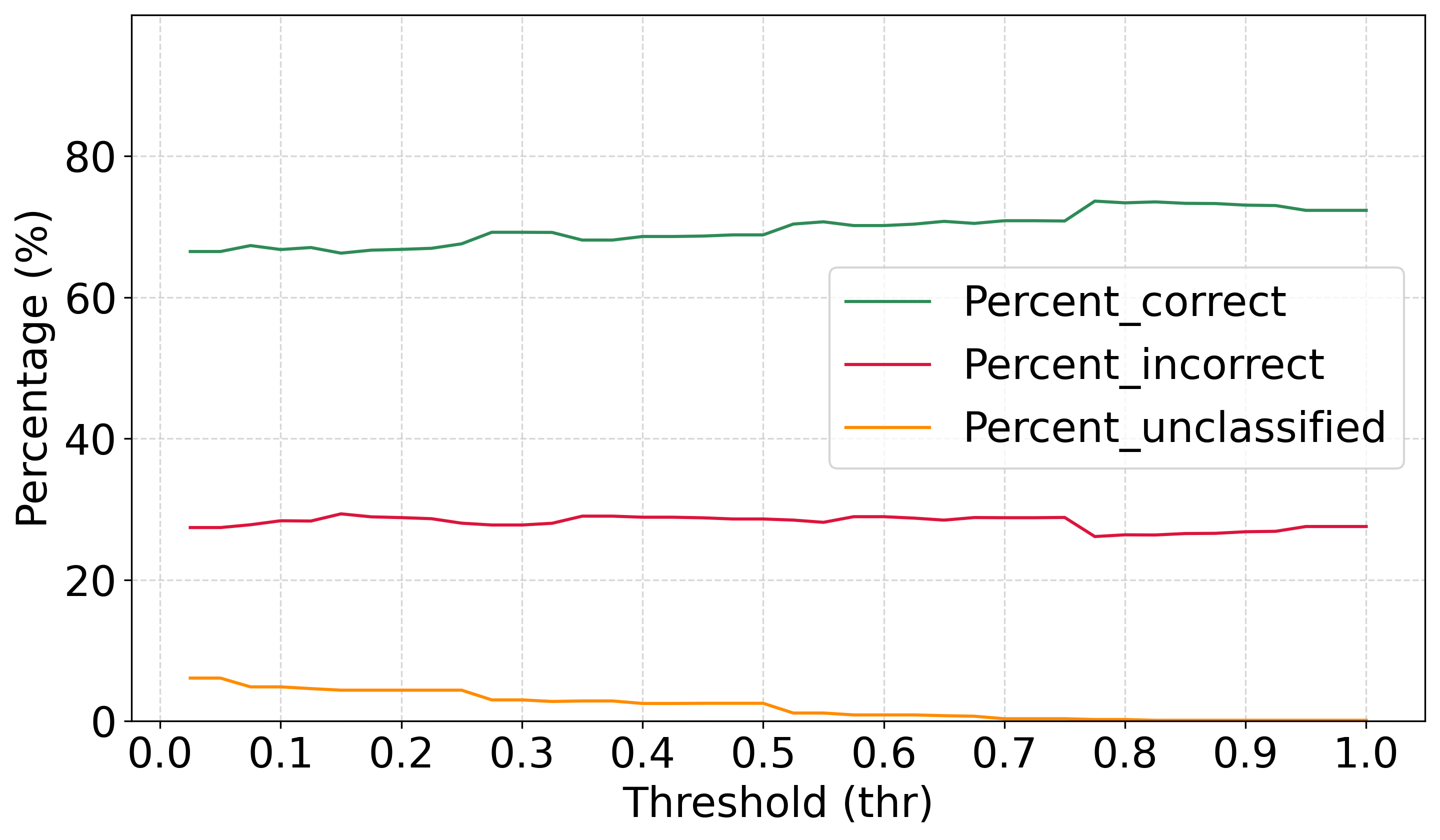}
\caption{CNC-TP with all values: evolution of correct, incorrect, and unclassified instance rates as a function of the threshold}
\label{figure:CNC-above-threshold-all-values-percents}
\end{figure}

\par As shown in Figure~\ref{figure:CNC-above-threshold-all-values-percents}, the percentage of correctly classified instances increases steadily with the threshold, reaching a peak of \textbf{73.642\%} at a threshold of \textbf{0.775}. 

In contrast, the rates of incorrectly classified and unclassified instances both decrease as the threshold increases, reaching their minimum values at the same threshold. 
This indicates that a threshold of 0.775 offers an optimal trade-off between classification accuracy and rejection rate, maximizing the classifier’s effectiveness while minimizing uncertainty.

\begin{figure}[!ht]
\centering
\includegraphics[width=1\linewidth]{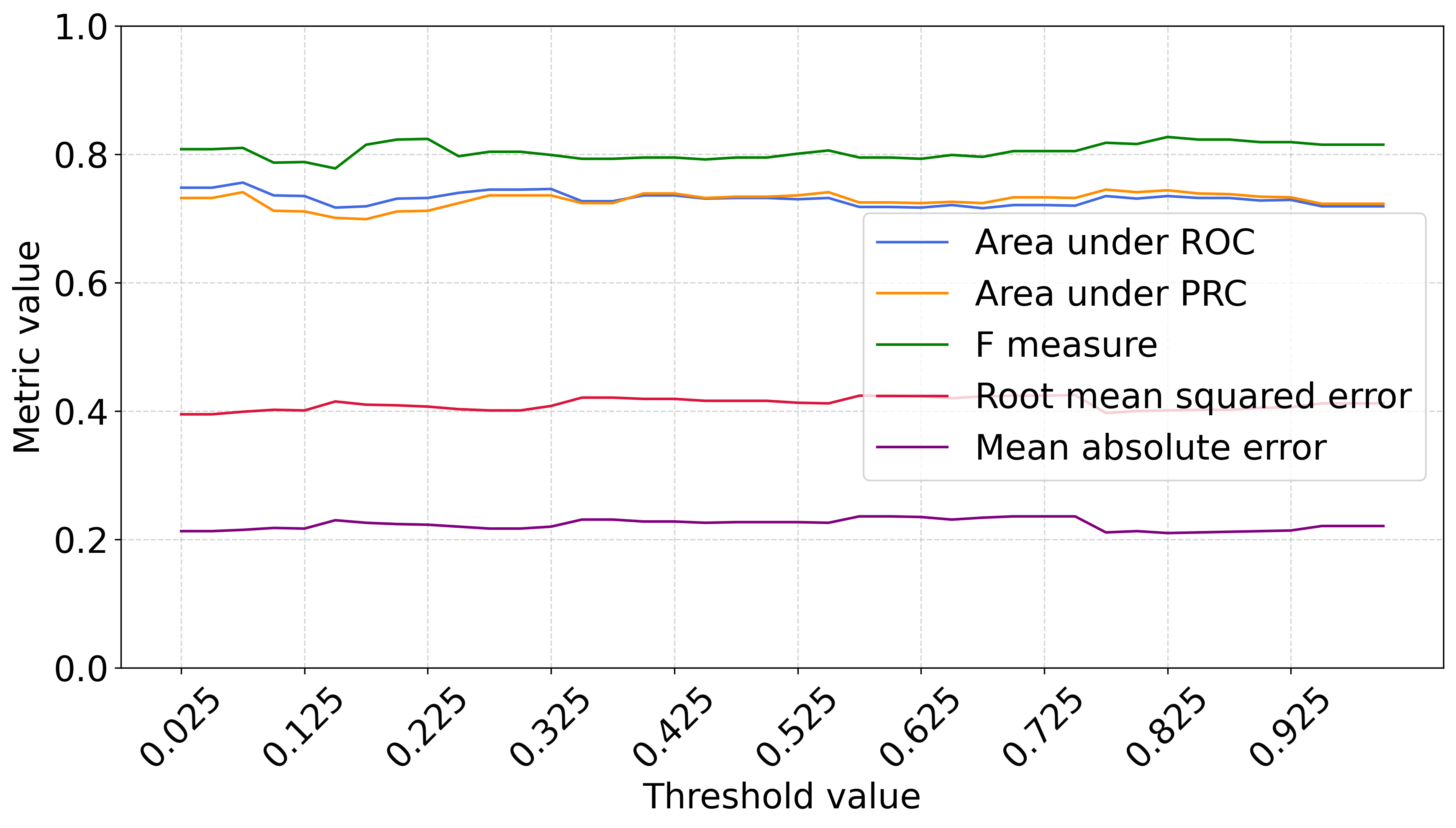}
\caption{CNC-TP with all values: evolution of additional performance metrics as a function of the threshold}
\label{figure:CNC-above-threshold-all-values-metrics}
\end{figure}

\par Figure~\ref{figure:CNC-above-threshold-all-values-metrics} illustrates the performance of the \textbf{CNC-TP} classifier (variant CNC-TP-AV) using several evaluation metrics as the attribute selection threshold varies.
The evolution of the \textbf{F1-score} is nearly linear and peaks around a threshold of 0.775, indicating a strong balance between precision and recall. This suggests that the classifier is not only accurate but also consistent in identifying both positive and negative instances.
The \textbf{Area Under the ROC Curve (AUC-ROC)} value reaches 0.735 and the \textbf{Area Under the Precision-Recall Curve (AUC-PRC)} 0.745.
These values confirm that CNC-TP-AV is a reliable learning model capable of effectively distinguishing between classes, including minority ones.
Regarding error metrics, the \textbf{Root Mean Squared Error (RMSE)} reaches a value of 0.397, while the \textbf{Mean Absolute Error (MAE)} reaches 0.211. 
These results indicate that the model tends to minimize large prediction errors (RMSE) and maintains a low average error (MAE), reflecting both robustness and stability.

\par When comparing Figures~\ref{figure:CNC-above-threshold-all-values-percents} and~\ref{figure:CNC-above-threshold-all-values-metrics}, a notable performance shift is observed around the threshold value of \textbf{0.775}. 
This threshold corresponds to the highest classification accuracy and the best trade-off between precision and recall, making it a strong candidate for the optimal configuration of the CNC-TP classifier.

\subsection{CNC-TP-RV: Top Pertinent Attributes with Relevant Values}
\label{subsection:CNC-TP-RV}

\par As a reminder, an alternative variant of the CNC-TP method is also possible. 
In this configuration, instead of computing the closure for all nominal values of a selected attribute, we focus only on the most relevant/frequent value. 
This strategy aims to reduce computational complexity while preserving classification effectiveness.
In line with the previous experiment, we vary the attribute selection threshold from 0.025 (2.5\%) to 1.000 (100\%) to identify the optimal value that yields the best performance.

\begin{figure}[!ht]
\centering
\includegraphics[width=1\linewidth]{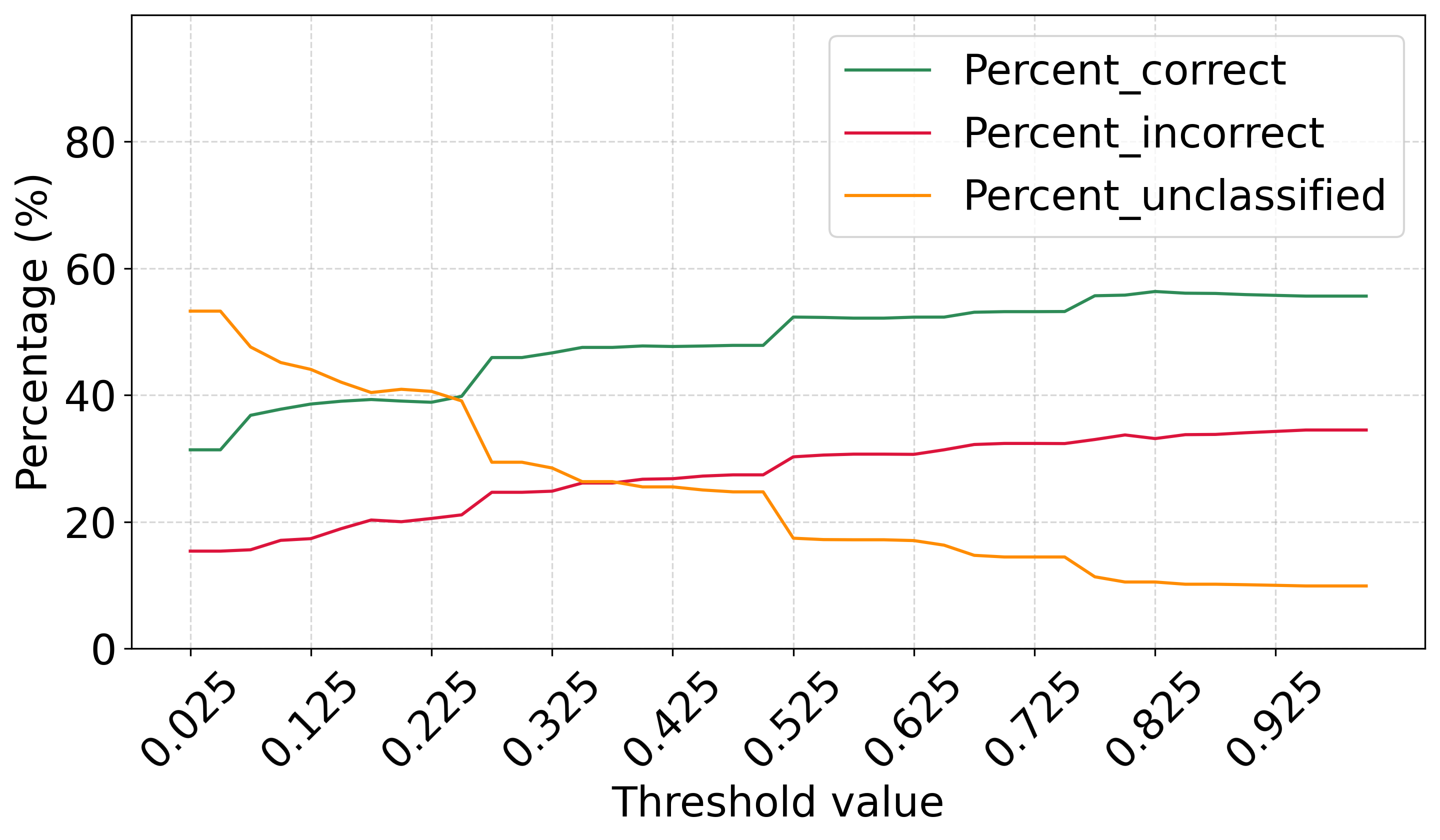}
\caption{CNC-TP with relevant values: evolution of correct, incorrect, and unclassified instance rates}
\label{figure:CNC-above-threshold-majority-values-percents}
\end{figure}

\par Figure~\ref{figure:CNC-above-threshold-majority-values-percents} shows the evolution of classification performance for the \textbf{CNC-TP-RV} variant using the second strategy described in Section~\ref{subsection:Concept_Extraction_Strategies}. 
As the threshold increases, the percentage of correctly classified instances rises, reaching a maximum of \textbf{55.684\%} at a threshold of \textbf{0.775}. Similarly, the rate of incorrectly classified instances peaks at \textbf{32.993\%} at the same threshold.
Notably, this variant achieves the lowest rejection rate (unclassified instances) at the threshold of 0.775, indicating that most instances are covered by at least one rule. 
However, the overall performance of CNC-TP-RV remains significantly lower than that of the previous variant (CNC-TP-AV), both in terms of accuracy and balance between precision and recall.
These results suggest that while CNC-TP-RV offers computational advantages by simplifying concept generation, it may sacrifice classification quality. 
Therefore, its use should be considered in contexts where interpretability and efficiency are prioritized over predictive accuracy.

\begin{figure}[!ht]
\centering
\includegraphics[width=1\linewidth]{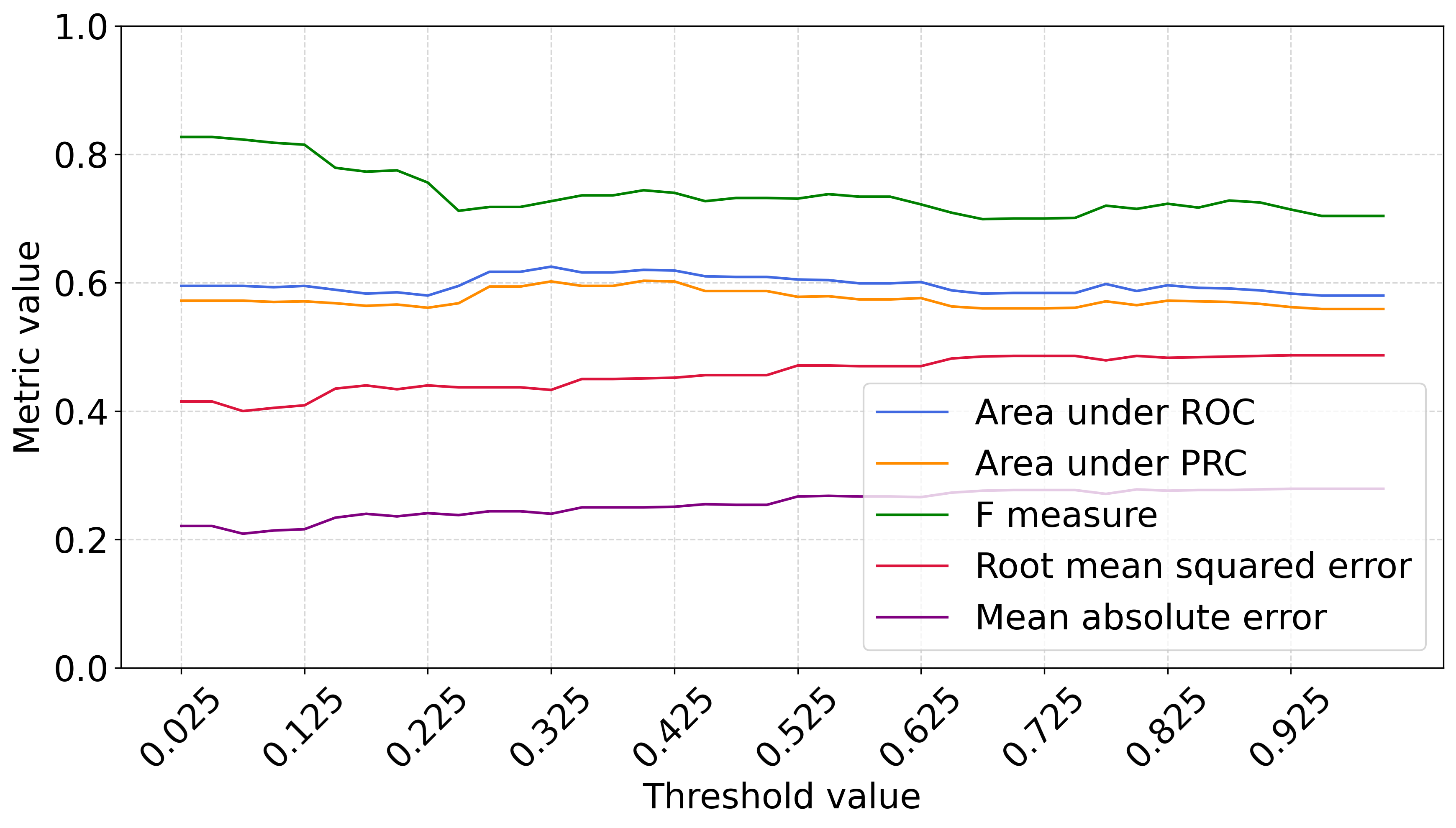}
\caption{CNC-TP with relevant values: evolution of additional performance metrics}
\label{figure:CNC-above-threshold-majority-values-metrics}
\end{figure}

\par Figure~\ref{figure:CNC-above-threshold-majority-values-metrics} presents the evolution of various performance metrics for the \textbf{CNC-TP-RV} variant, as the attribute selection threshold increases.
The \textbf{F1-score} shows a decreasing trend, dropping from 0.823 to 0.720. 
This indicates that the model struggles to maintain a good balance between precision and recall as the threshold increases, suggesting a degradation in classification quality.
Similarly, both the \textbf{Area Under the ROC Curve (AUC-ROC)} and the \textbf{Area Under the Precision-Recall Curve (AUC-PRC)} decline, ranging between 0.560 and 0.603. 
These values are relatively low, and the proximity of AUC-ROC to 0.5 suggests that the model performs only marginally better than random guessing in some configurations.
The \textbf{Root Mean Squared Error (RMSE)} varies between 0.40 and 0.487, while the \textbf{Mean Absolute Error (MAE)} ranges from 0.209 to 0.279. 
These results indicate that although the model maintains a moderate level of error control, it does not compensate for the overall drop in classification performance.

\par Comparing Figures~\ref{figure:CNC-above-threshold-all-values-percents} and~\ref{figure:CNC-above-threshold-all-values-metrics}, we observe that performance metrics tend to stabilize once the threshold exceeds 0.775. 
However, the overall performance of this variant remains inferior to the one presented in the previous section.
Based on these findings, we choose to continue the experimental evaluation using the CNC-TP variant that computes closures for \textbf{All Values} of attributes whose pertinence exceeds the threshold of 0.775, as it consistently delivers better classification results.

\subsection{CNC-TP Compared to Other Classifiers}
\label{subsection:CN-TP_Compared_other_classifiers}

\par Since the proposed CNC-TP method generates classification rules, we begin by comparing it with several well-known rule-based classifiers from the literature. 
These include: \textit{ConjunctiveRule}~\cite{xu2014}, \textit{DTNB}~\cite{ferreira2025double}, \textit{DecisionTable}~\cite{comp_class_2019}, \textit{FURIA}~\cite{zhang2024belief}, \textit{JRip}~\cite{kalmegh2024hypothyroid}, \textit{NNge}~\cite{SIFO2024}, \textit{OLM}~\cite{Lee2022}, \textit{OneR}~\cite{Zhang2023}, \textit{PART}~\cite{Tran2024}, \textit{Ridor}~\cite{Najadat2020}, \textit{RoughSet}~\cite{Li2023}, and \textit{ZeroR}~\cite{Sangeorzan2023}.

\begin{figure}[!ht]
\centering
\includegraphics[width=1\linewidth]{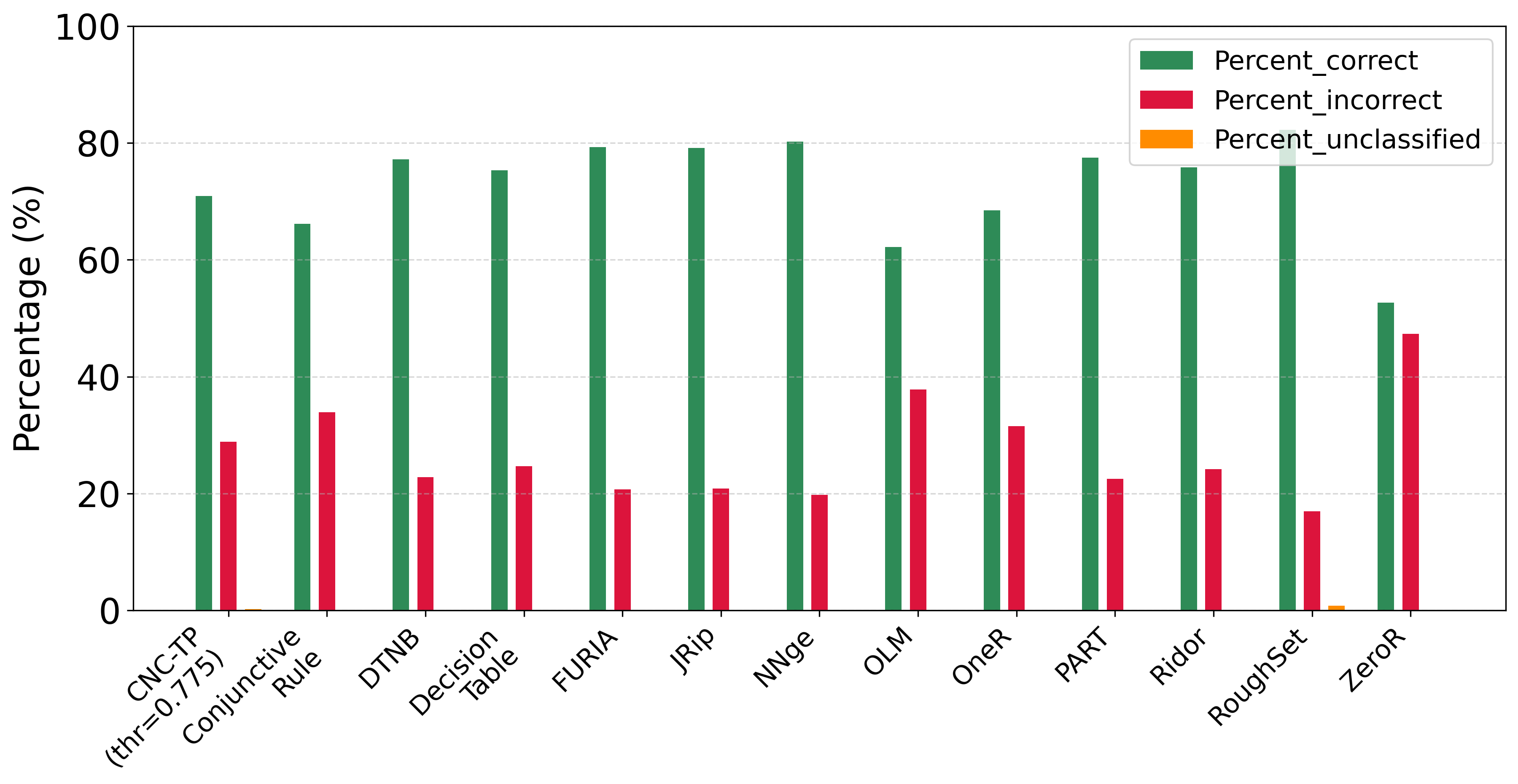}
\caption{CNC-TP compared to rule-based methods: evolution of correct, incorrect, and unclassified instance rates}
\label{figure:CNC-vs-Rules-percents}
\end{figure}

\par As shown in Figure~\ref{figure:CNC-vs-Rules-percents}, CNC-TP demonstrates a lower rejection rate compared to \textit{RoughSet}, and outperforms \textit{ConjunctiveRule}, \textit{OLM}, \textit{OneR}, and \textit{ZeroR} in terms of classification accuracy. 
Furthermore, its performance is comparable to more advanced rule-based classifiers such as \textit{DTNB}, \textit{DecisionTable}, \textit{FURIA}, \textit{NNge}, \textit{PART}, and \textit{Ridor}.

\begin{figure}[!ht]
\centering
\includegraphics[width=1\linewidth]{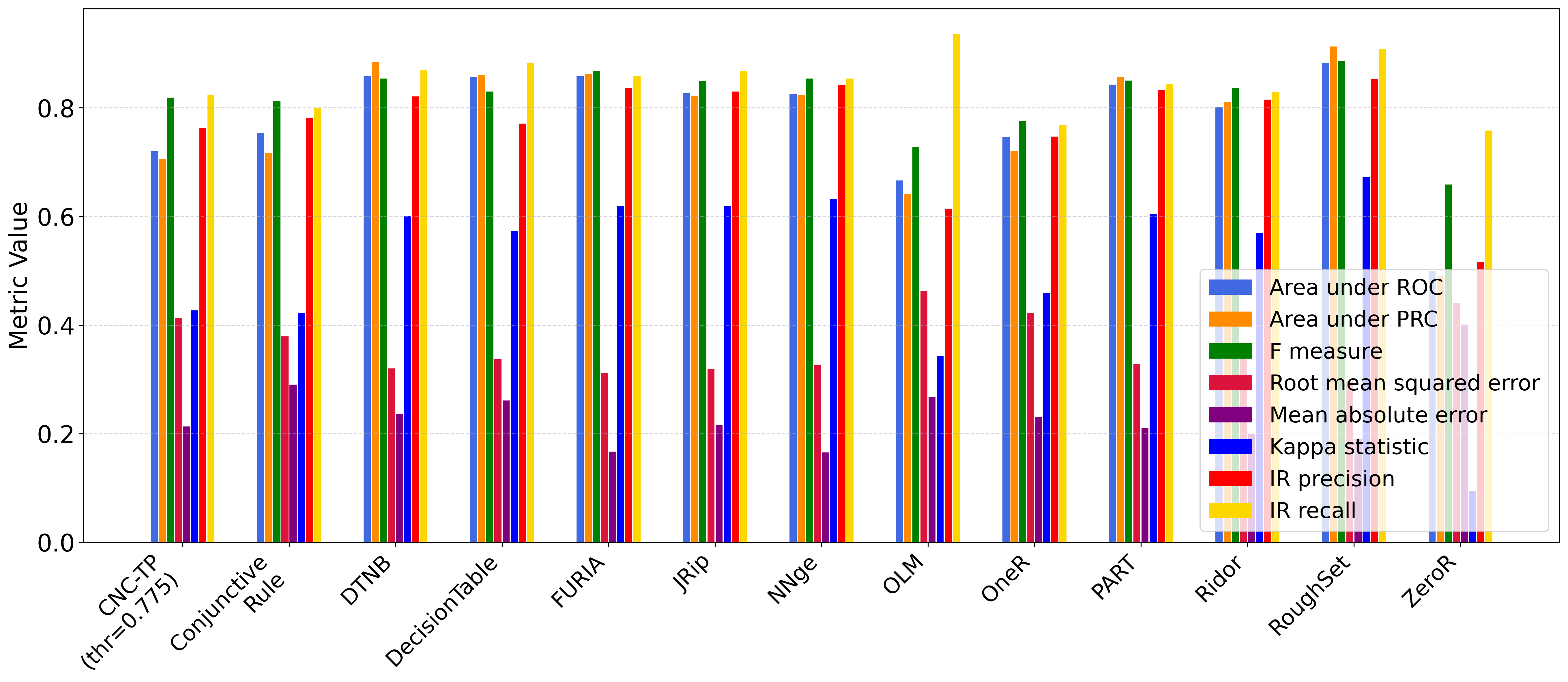}
\caption{CNC-TP compared to rule-based methods: evolution of F-measure, precision, and recall}
\label{figure:CNC-vs-Rules-metrics}
\end{figure}

\par Figure~\ref{figure:CNC-vs-Rules-metrics} further highlights the strengths of CNC-TP. It achieves one of the highest \textbf{F1-scores}, close to 1, indicating an excellent balance between precision and recall. 
Both precision and recall values are around 0.8, suggesting that CNC-TP produces few false positives while successfully identifying most positive instances.

\par Regarding the \textbf{Kappa statistic}, CNC-TP ranks in the mid-range among the compared classifiers. 
While \textit{RoughSet} achieves the highest Kappa value (0.673), and \textit{OLM} the lowest (0.343), CNC-TP maintains a solid agreement between predictions and actual labels. 
Notably, \textit{ZeroR} stands out with a near-zero Kappa, reflecting its inability to capture any meaningful classification pattern.

\par Overall, CNC-TP proves to be a competitive rule-based classifier, offering a strong compromise between interpretability, accuracy, and robustness. 
Its performance is on par with or superior to several established methods, making it a viable alternative for interpretable machine learning tasks.

\begin{figure}
\centering
\includegraphics[width=1\linewidth]{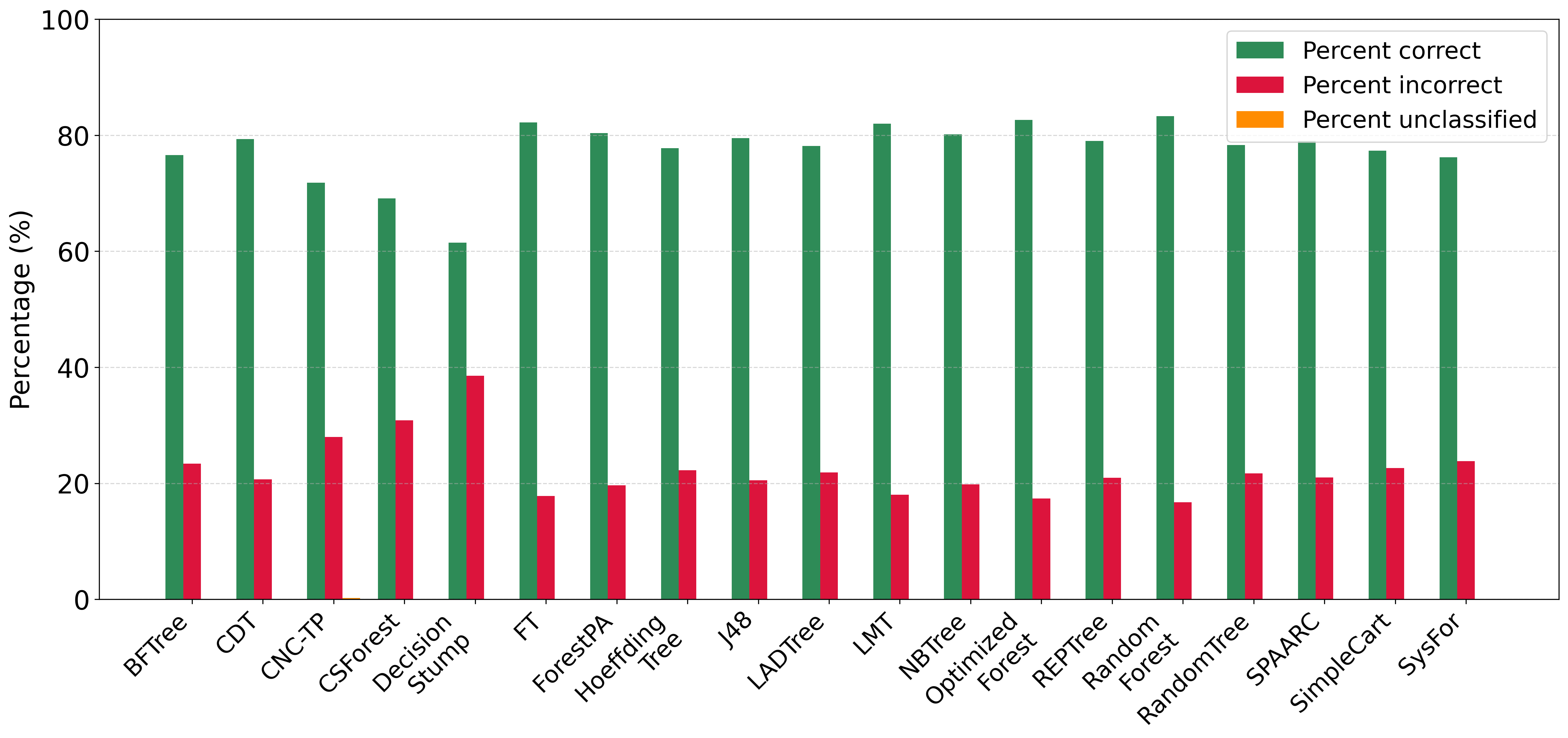}
\caption{CNC-TP compared to tree-based methods: evolution of correct, incorrect, and unclassified instance rates}
\label{figure:CNC-vs-Tree-percents}
\end{figure}

\begin{figure}
\centering
\includegraphics[width=1\linewidth]{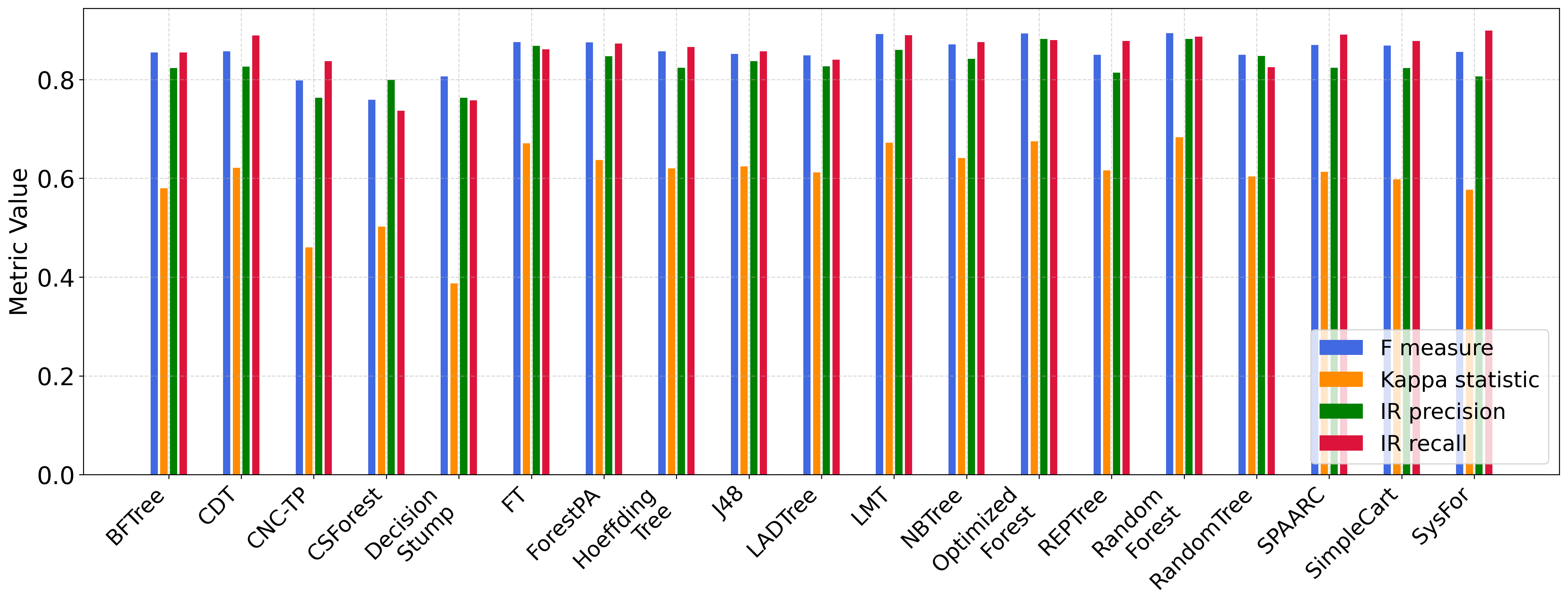}
\caption{CNC-TP compared to tree-based methods: evolution of F-measure, precision, and recall}
\label{figure:CNC-vs-Tree-metrics}
\end{figure}

\par In the continuation of our experiments, we compare \textbf{CNC-TP} with a set of interpretable and explainable classifiers based on decision trees. 
These include: BFTree, CDT, CSForest, DecisionStump, FT, ForestPA, HoeffdingTree, J48, LADTree, LMT, NBTree, OptimizedForest, REPTree, RandomForest, RandomTree, SPAARC, SimpleCart, and SysFor.
According to Figures~\ref{figure:CNC-vs-Tree-percents} and~\ref{figure:CNC-vs-Tree-metrics}, CNC-TP demonstrates globally competitive performance. 
It achieves classification accuracy and F-measure values comparable to, and in some cases better than, several tree-based methods such as \textit{CSForest} and \textit{DecisionStump}. 
Moreover, CNC-TP maintains a relatively low rejection rate while preserving a strong balance between precision and recall.

\par These results confirm that CNC-TP is not only interpretable and rule-based, but also competitive with state-of-the-art decision tree classifiers in terms of predictive performance. 
This makes it a promising alternative for applications requiring both transparency and accuracy.

\begin{figure}[!ht]
\centering
\includegraphics[width=1\linewidth]{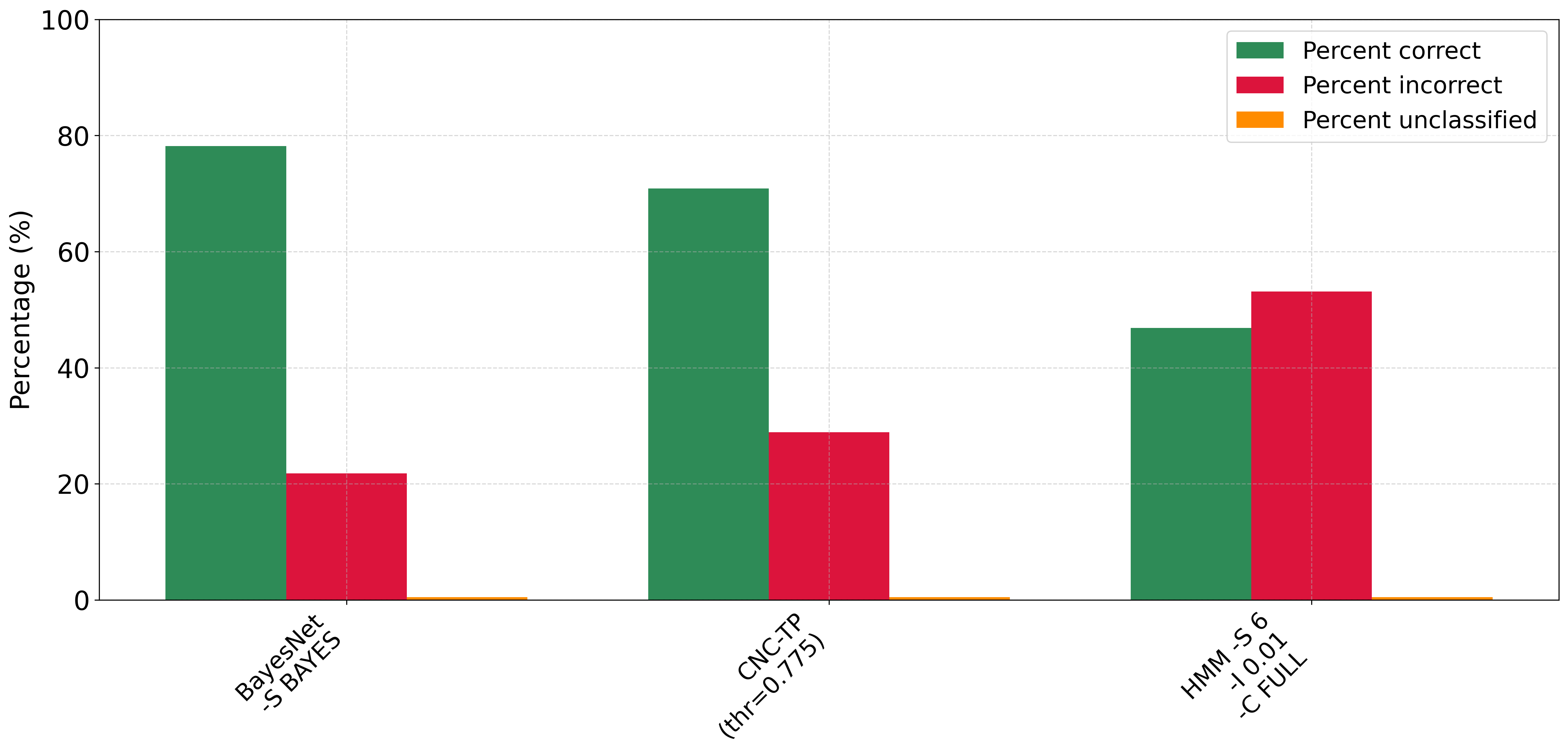}
\caption{CNC-TP compared to Bayesian methods: evolution of correct, incorrect, and unclassified instance rates}
\label{figure:CNC-vs-Bayesians-percents}
\end{figure}

\begin{figure}[!ht]
\centering
\includegraphics[width=1\linewidth]{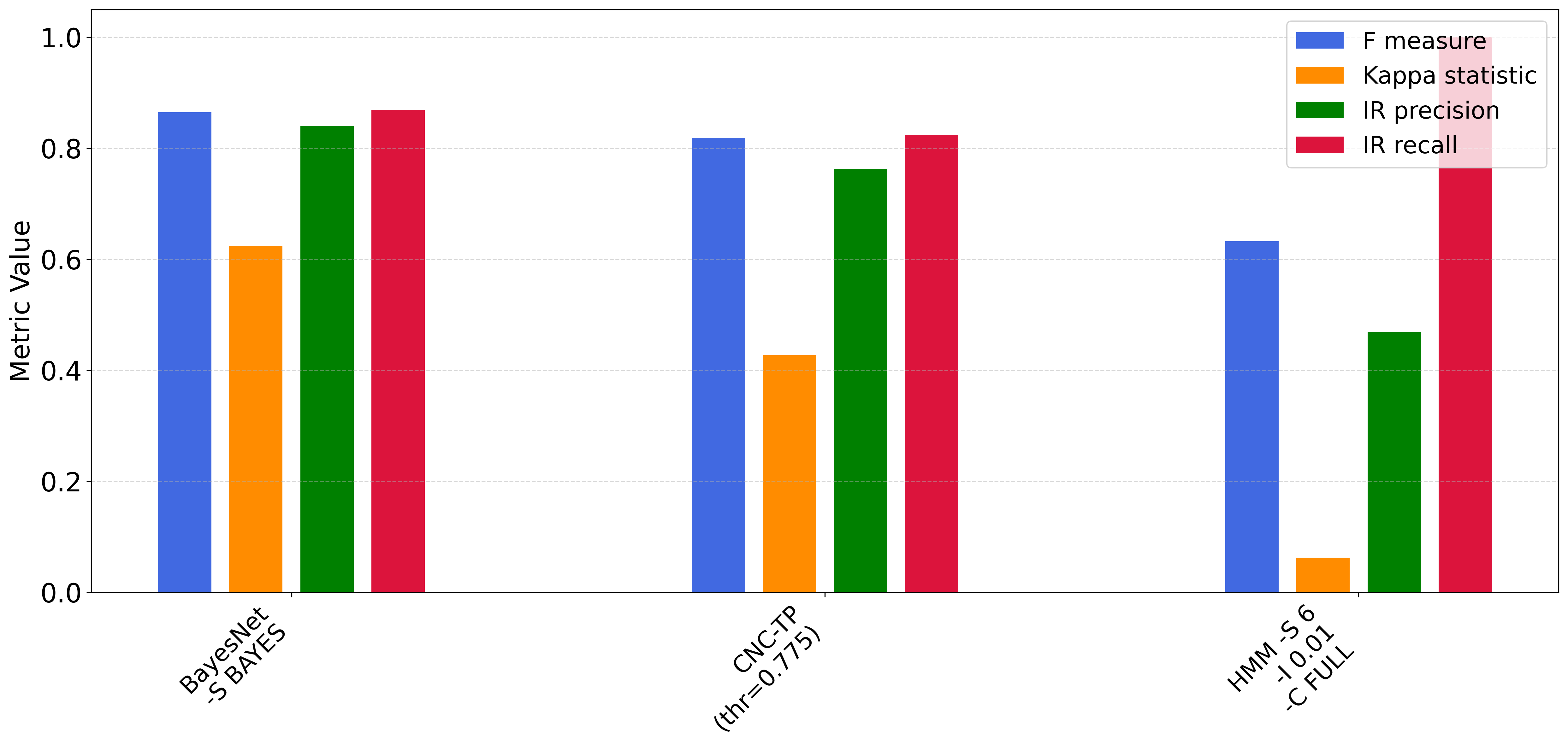}
\caption{CNC-TP compared to Bayesian methods: evolution of F-measure, precision, and recall}
\label{figure:CNC-vs-Bayesians-metrics}
\end{figure}

\par In the next part of our experimental study, we compare \textbf{CNC-TP} with two well-known Bayesian classifiers: \textit{BayesNet}~\cite{Pearl1988,Xi2024} and \textit{HMM (Hidden Markov Model)}~\cite{Baum1966,Williams2024}.
As shown in Figure~\ref{figure:CNC-vs-Bayesians-percents}, CNC-TP achieves a higher classification accuracy than HMM and performs comparably to BayesNet. 
It also maintains a lower rejection rate than both Bayesian methods, indicating better rule coverage and generalization.
Figure~\ref{figure:CNC-vs-Bayesians-metrics} provides further insights. CNC-TP achieves precision, recall, and F1-score values close to those of BayesNet, confirming its ability to balance false positives and false negatives effectively. 
However, CNC-TP shows slightly lower performance in terms of the \textbf{Kappa statistic}, suggesting a slightly weaker agreement between predicted and actual labels compared to BayesNet.
HMM, on the other hand, exhibits relatively low precision, F1-score, and Kappa values, despite achieving a higher recall. 
This indicates that while HMM is sensitive to positive instances, it tends to produce more false positives and lacks overall predictive reliability.
In summary, CNC-TP proves to be a competitive alternative to probabilistic classifiers, offering comparable performance to BayesNet while maintaining the advantages of interpretability and rule-based reasoning.

\begin{figure}
\centering
\includegraphics[width=1\linewidth]{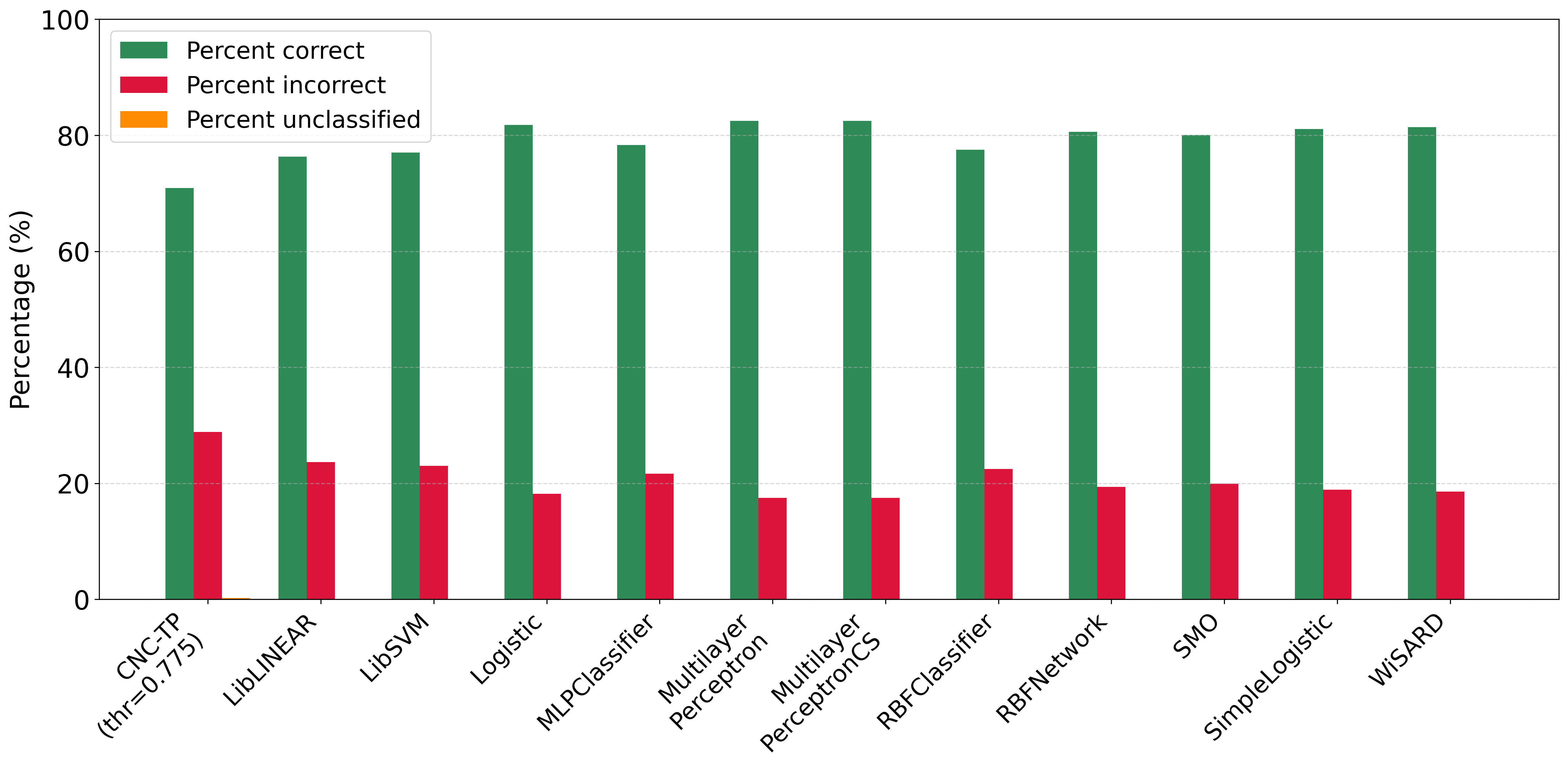}
\caption{CNC-TP compared to statistical machine learning methods: evolution of correct, incorrect, and unclassified instance rates}
\label{figure:CNC-vs-MLstats-percents}
\end{figure}
\begin{figure}
\centering
\includegraphics[width=1\linewidth]{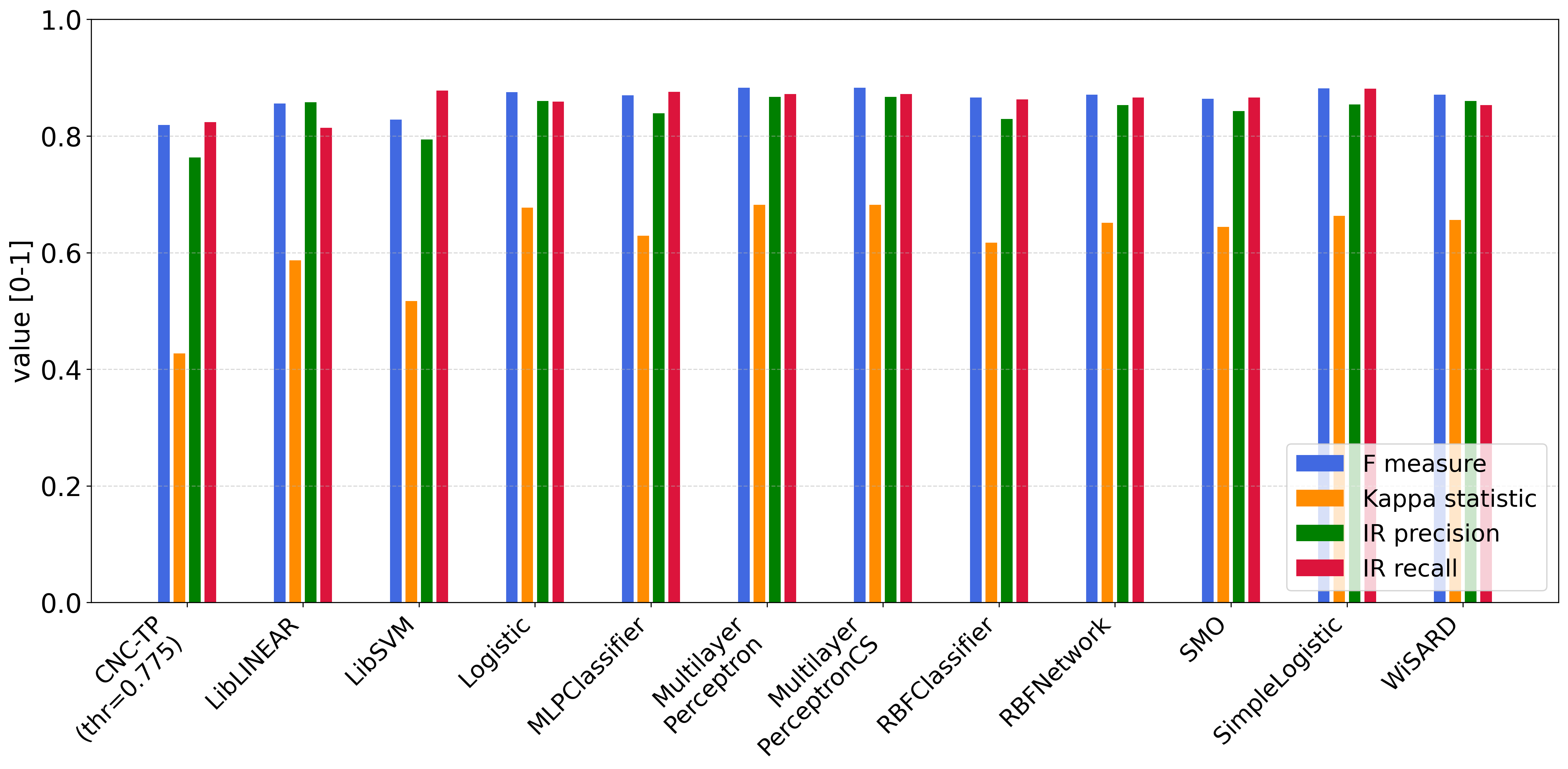}
\caption{CNC-TP compared to statistical machine learning methods: evolution of F-measure, precision, and recall}
\label{figure:CNC-vs-MLstats-metrics}
\end{figure}
\par We also compare \textbf{CNC-TP} with a set of machine learning methods grounded in statistical modeling.
These include: LibLINEAR, LibSVM, Logistic Regression, MLPClassifier, MultilayerPerceptron, MultilayerPerceptronCS, RBFClassifier, SMO, SimpleLogistic, and WiSARD.
According to Figures~\ref{figure:CNC-vs-MLstats-percents} and~\ref{figure:CNC-vs-MLstats-metrics}, CNC-TP consistently demonstrates performance that is comparable to these statistical classifiers. 
It achieves competitive accuracy, precision, and recall, while maintaining a low rejection rate. 
Although some methods such as \textit{LibSVM} or \textit{MLPClassifier} may slightly outperform CNC-TP in certain metrics, the proposed method offers the added advantage of interpretability and rule-based transparency, which are often lacking in black-box statistical models.
These results confirm that CNC-TP is a viable alternative to traditional statistical learning techniques, particularly in contexts where explainability and decision traceability are essential.


\section{Conclusion}
\label{section:conclusion}

\par In this study, we introduced \textbf{CNC-TP}, a novel classification approach based on Formal Concept Analysis (FCA), which integrates attribute selection using the Gain Ratio measure with rule-based learning. 
Two concept extraction strategies were explored: one using all values of the top-ranked attributes, and another using only their most frequent (relevant) values.

\par Experimental results across 16 diverse datasets demonstrated that the variant using \textbf{all values} of the top-$p\%$ attributes (with $p = 0.775$) consistently outperformed the majority-value variant in terms of accuracy, F1-score, and error minimization. 
This configuration also achieved a strong balance between precision and recall, while maintaining a low rejection rate.

\par When compared to a wide range of classifiers—including rule-based learners, decision trees, Bayesian models, and statistical machine learning methods CNC-TP showed competitive performance. 
It outperformed several interpretable models such as \textit{ConjunctiveRule}, \textit{OneR}, and \textit{DecisionStump}, and achieved results comparable to more advanced classifiers like \textit{FURIA}, \textit{J48}, \textit{BayesNet}, and \textit{LibSVM}. 
While some models slightly surpassed CNC-TP in specific metrics (e.g., Kappa statistic), CNC-TP maintained the advantage of interpretability and explainability through its rule-based structure.

\par Overall, CNC-TP offers a promising trade-off between performance and interpretability. 
Its ability to adapt to different data characteristics, combined with its transparent decision-making process, makes it a valuable tool for domains where explainability is essential. 
Future work may explore hybrid strategies that combine both concept extraction variants or integrate fuzzy logic to further enhance classification robustness.


\bibliographystyle{IEEEtran}
\bibliography{bibliography}

\end{document}